\documentclass[10pt,twocolumn,letterpaper]{article}

\usepackage{cvpr}              

\usepackage{graphicx}
\usepackage{amsmath}
\usepackage{amssymb}
\usepackage{booktabs}

\usepackage[pagebackref,breaklinks,colorlinks]{hyperref}

\usepackage[capitalize]{cleveref}
\crefname{section}{Sec.}{Secs.}
\Crefname{section}{Section}{Sections}
\Crefname{table}{Table}{Tables}
\crefname{table}{Tab.}{Tabs.}

\def\cvprPaperID{9796} 
\def\confName{CVPR}
\def\confYear{2022}

\begin{document}

\title{DECORE: Deep Compression with Reinforcement Learning}

\author{Manoj Alwani\\
Element Inc.\\
{\tt\small ma@discoverelement.com}
\and
Yang Wang\\
Element Inc.\\
{\tt\small yw@discoverelement.com}
\and
Vashisht Madhavan\\
Element Inc.\\
{\tt\small vm@discoverelement.com}
}
\maketitle

\begin{abstract}
   Deep learning has become an increasingly popular and powerful methodology for modern pattern recognition systems. However, many deep neural networks have millions or billions of parameters, making them untenable for real-world applications due to constraints on memory size or latency requirements. As a result, efficient network compression techniques are often required for the widespread adoption of deep learning methods. We present DECORE, a reinforcement learning-based approach to automate the network compression process. DECORE assigns an agent to each channel in the network along with a light policy gradient method to learn which neurons or channels to be kept or removed. Each agent in the network has just one parameter (keep or drop) to learn, which leads to a much faster training process compared to existing approaches. DECORE provides state-of-the-art compression results on various network architectures and various datasets. For example, on the ResNet-110 architecture, it achieves a 64.8\% compression and 61.8\% FLOPs reduction as compared to the baseline model without any accuracy loss on the CIFAR-10 dataset. It can reduce the size of regular architectures like the VGG network by up to 99\% with just a small accuracy drop of 2.28\%. For a larger dataset like ImageNet with just 30 epochs of training, it can compress the ResNet-50 architecture by 44.7\% and reduce FLOPs by 42.3\%,  with just a 0.69\% drop on Top-5 accuracy of the uncompressed model. We also demonstrate that DECORE can be used to search for compressed network architectures based on various constraints, such as memory and FLOPs.
\end{abstract}

\section{Introduction}
\label{sec:intro}
Deep neural networks (DNNs) have led to significant advances in image recognition tasks, most notably benchmarked by the ImageNet challenge~\cite{imagenet}. Recent trends suggest that deeper architectures with more parameters and more complex computational blocks lead to better results~\cite{simonyan2014, resnet2015}. These increases in model size, however, incur a higher inference cost, making them difficult to be deployed in embedded systems or mobile devices. Additionally, real-world applications have memory, latency, or throughput constraints which make it difficult to use large convolutional neural networks (CNNs) as it is. Therefore, lowering the inference time and memory consumption of large models would benefit the wider adoption of deep learning models for these real-world applications. 

There has been a significant amount of work on reducing the inference cost via a more efficient network architecture~\cite{MobileNetv2, morphnet, li2016pruning, DentonZBLF14}. Although much progress has been made, the process of finding efficient architectures often requires a lot of experience and manual design. The automation of this process is an active research area of great value, as newer, more powerful DNN architectures are rapidly being discovered. Meanwhile, a large body of work has also been done in direction of network compression, such as low-rank decomposition~\cite{DentonZBLF14,zhang2019cvpr, Lin2016IJCAI, Lin2018PAMI}, parameter quantization~\cite{jacob2018cvpr, bohan2018cvpr, rastegeri2018xnornet}, and pruning~\cite{li2016pruning, gdp2018, lin2020hrank, yu2018nisp, lin2019gal, Hu2016arxiv}. 

These compression techniques can be broadly divided into two categories: non-structured and structured pruning. Non-structured pruning methods~\cite{han2015deep, han2015learning, hassibi1993optimal, babak1993} directly prune weights of CNN to get sparse weight matrices. Along with compression, these approaches can provide faster network inference using specialized software~\cite{han2016eie} or hardware~\cite{jongsoo2017iclr}. However, they often cause irregular memory access that adversely affects the network inference on general-purpose hardware or BLAS libraries. Structured pruning overcomes these limitations by removing structured weights like 2D kernels and feature maps (channels) or layers, which focus on network architecture changes and can be supported by off-the-shelf deep learning libraries like Pytorch~\cite{PyTorch}, and TensorFlow~\cite{AbaBar16Tensorflow}  etc. Pruning structured weights is a complex problem as we have to find out which layers or channels are least important in the network and removing them will not affect the accuracy significantly. For example, in VGG16~\cite{simonyan2014} the first hidden layer has 64 channels, an exhaustive search for which channels to remove without affecting accuracy involves $2^{64}$ combinations. For the whole network with around $5,000$ channels in total, it is computationally infeasible to go through all combinations. To overcome this problem many techniques have been proposed in the following categories.

\textbf{Pruning using trained network statistics:} These approaches mostly consist of two steps, first pruning the weights using network statistics and then fine-tuning the compressed network to recover from accuracy loss. Among these techniques, magnitude-based pruning has been proposed to find filters~\cite{li2016pruning} or channels (feature maps)~\cite{Hu2016arxiv} with the lowest $l1$-norm as they are least informative and can be removed. Yang et al.~\cite{yang2019cvpr} calculates the geometric median of each layer and prunes the filters close to it. Greedy pruning~\cite{He2017ICCV, thinet2017} is also a common strategy that prunes filters by considering statistics computed from the next layer in a greedy layer-wise manner. More recently, Lin et al.~\cite{lin2020hrank} proposed to find the channel's importance by calculating the rank of channels. All these methods calculate channel or filter importance at one or neighboring layers without considering all other layers in a deep neural network. Yu et al.~\cite{yu2018nisp} found that channels that were assumed less important and pruned in early layers could have a significant impact on the pruning of later layers. To address this issue, they calculate the importance score of the final layer and propagate it to each filter in the network to get its importance score. 

Finding the least important channels using trained network statistics is fairly straightforward, but finding how many channels or filters to drop from each layer (or the network) to achieve a certain compression rate requires manual effort. For example, in~\cite{lin2020hrank, yu2018nisp} a fixed compression rate is used before fine-tuning which limits exploration for different compression rates, as each combination requires fine-tuning to recover from accuracy loss which could be time-consuming.

\textbf{Pruning by learning channel importance:} Another group of techniques is learning which channels are important by applying compression constraints while retraining the network, such as sparse scaling parameters in batch normalization layers~\cite{tan2019efficientnet, zhao2019variational, liu2017learning}. Extension to these approaches has also taken into account of resource constraints such as latency and computation~\cite{morphnet, molchanov2016pruning, galvan2021neuroevolution}. Huang et al.~\cite{huang2018data} and Lin et al.~\cite{lin2019gal} introduced a new binary parameter to each channel (mask) to find out if the channel is important or not, but learning this binary mask involves an NP-hard optimization. This joint optimization for both compression and accuracy leads to higher accuracy performance but since the (combined) loss function is different than the original network, it requires specialized optimizers~\cite{lin2019gal, huang2018data} and needs iterative fine-tuning which increases time for both model training and manual adjustments.

\textbf{Pruning by architecture search:} Given that designing deep network architectures requires significant human effort, there have been some explorations on automatically learning network architectures. These approaches build networks from the ground up using a set of custom building blocks, relying on variations of a trial and error search to find promising architectures~\cite{tan2019mnasnet, tan2019efficientnet, liu2018darts, jin2016submodular, zoph2016neural}, or compressed ones. In~\cite{AMC_He_2018_ECCV, ashok2018nn}, Reinforcement Learning (RL) is used to find compressed architectures but these techniques often take a long time to train. For example, in~\cite{ashok2018nn} a knowledge distillation~\cite{hinton2015distilling, ba2013deep} approach is used to find a compressed student network from a teacher network using RL. The method utilizes the architecture search to reduce the depth of a network and size of layers which can take up to 2,500 epochs. This search cost is however prohibitively expensive for regular deep learning practitioners and further exacerbated by the growth in the dataset and network complexities. To reduce the architecture search time, generative adversarial networks have been used in~\cite{lin2019gal} to find compressed structures. It shortens the search time significantly but still needs many iterations to train both generative and adversarial networks. Although these methods have led to a high compression rate, they often require a large amount of GPU time to obtain high-performing architectures.

Although the approaches mentioned above provide state-of-the-art results, they are computationally expensive and time-consuming as they require multiple iterations of pruning and fine-tuning, and often need a lot of manual effort. In this paper, we present DECORE, a multi-agent reinforcement learning (RL) framework for network compression which overcomes these limitations. In RL, an agent learns to perform a sequence of actions in an environment that maximizes some notion of a cumulative reward~\cite{sutton2018reinforcement}. In our approach, we assign dedicated agents to all channels in the network which take actions to drop or keep a channel in the network. The agent gets a positive reward when it drops the channel for higher compression, but it gets a negative reward (penalty) if the accuracy is decreased due to pruning. Using the REINFORCE algorithm~\cite{williams1992simple}, we optimize agents' policies to find out which channels to drop at each layer without affecting accuracy significantly and in turn maximizing the reward. Training of agents using the REINFORCE algorithm is independent of the network training loss which helps speed up the compression process. 

Our main contributions are summarized as follows:
\begin{itemize}
    \item We propose DECORE, a flexible and powerful approach to automating structure search and model compression. DECORE learns which channels are important by jointly training the network to provide high compression and FLOPs reduction rates.
    \item DECORE assigns an agent to each channel in the network (multi-agent learning) while each agent learns only one parameter, as opposed to other RL-based methods where the policy involves training another neural network. Learning a single parameter for each agent makes training much faster and more efficient.
    \item We demonstrate that our approach is able to find the most important channels in the network which can be used to search for compressed network architectures.
    \item Results on both CIFAR-10 and ImageNet datasets are reported with various networks to demonstrate that DECORE is able to achieve better accuracy, as well as higher compression and acceleration (FLOPs reduction) rates, compared to other existing methods.  
\end{itemize}

\section{Related Work}
\subsection{Network Pruning}
Pruning weights of DNNs has been an active research topic in deep architecture design and compression. As discussed in section~\ref{sec:intro}, weight pruning is effective at removing individual weights but does not necessarily reduce the computation time or memory usage. Sparse DNN layers can only help compress and accelerate the network with custom BLAS libraries or specialized hardware~\cite{han2016eie, iandola2016squeezenet}, which often have limited usage and an additional overhead~\cite{gray2017gpu, liu2015sparse}. A few approaches have also been proposed to leverage sparsity in networks by removing entire channels~\cite{li2016pruning, wen2016learning, scardapane2017group}.

There are other approaches to pruning channels by analyzing trained network statistics\cite{yuan2006model, simon2013sparse}, or learning the importance of each channel or filter in the network\cite{liu2017learning, huang2018data,lin2020hrank, morphnet}. Our approach is similar to these methods in terms of operating on entire channels, but using RL to identify important channels instead of relying on network statistics. Moreover, our approach learns channel importance while training the network without changing the loss function like in~\cite{lin2019gal, huang2018data}, which speeds up the overall training process.



\subsection{Architecture Search}
The design space of CNN architectures is extremely large and hence infeasible for an exhaustive manual search. Research on automating neural network design goes back to the 1980s when genetic algorithms were proposed to find both architectures and weights~\cite{Schaffer}. Modern architecture search approaches have led to state-of-the-art results for many tasks, but often require a significant amount of GPU time to obtain a high-performing architecture~\cite{zoph2016neural, tan2019mnasnet, tan2019efficientnet, liu2018darts, ashok2018nn, AMC_He_2018_ECCV}. Our use case is different, as we take already trained architectures and compress them while maintaining accuracy. This drastically reduces the search space and thus makes our approach more efficient.

Reinforcement learning has also been used to find compressed architectures from trained networks~\cite{AMC_He_2018_ECCV, ashok2018nn, zhan2019deep}. These approaches focus on finding a fixed set of custom rules for architecture design and then fine-tune the best architecture found. Although the result is a compressed network, these approaches still take a large amount of GPU hours to train. However, our proposed approach does not use any custom rules to look for compressed architectures. Like some of the regularization approaches, we simply learn which channels are more or less important for prediction. We obtain a compressed network by removing the channels that have little effect on accuracy.
\begin{figure}\label{fig:POI}
\centering
\includegraphics[width=8cm,height=10cm,keepaspectratio]{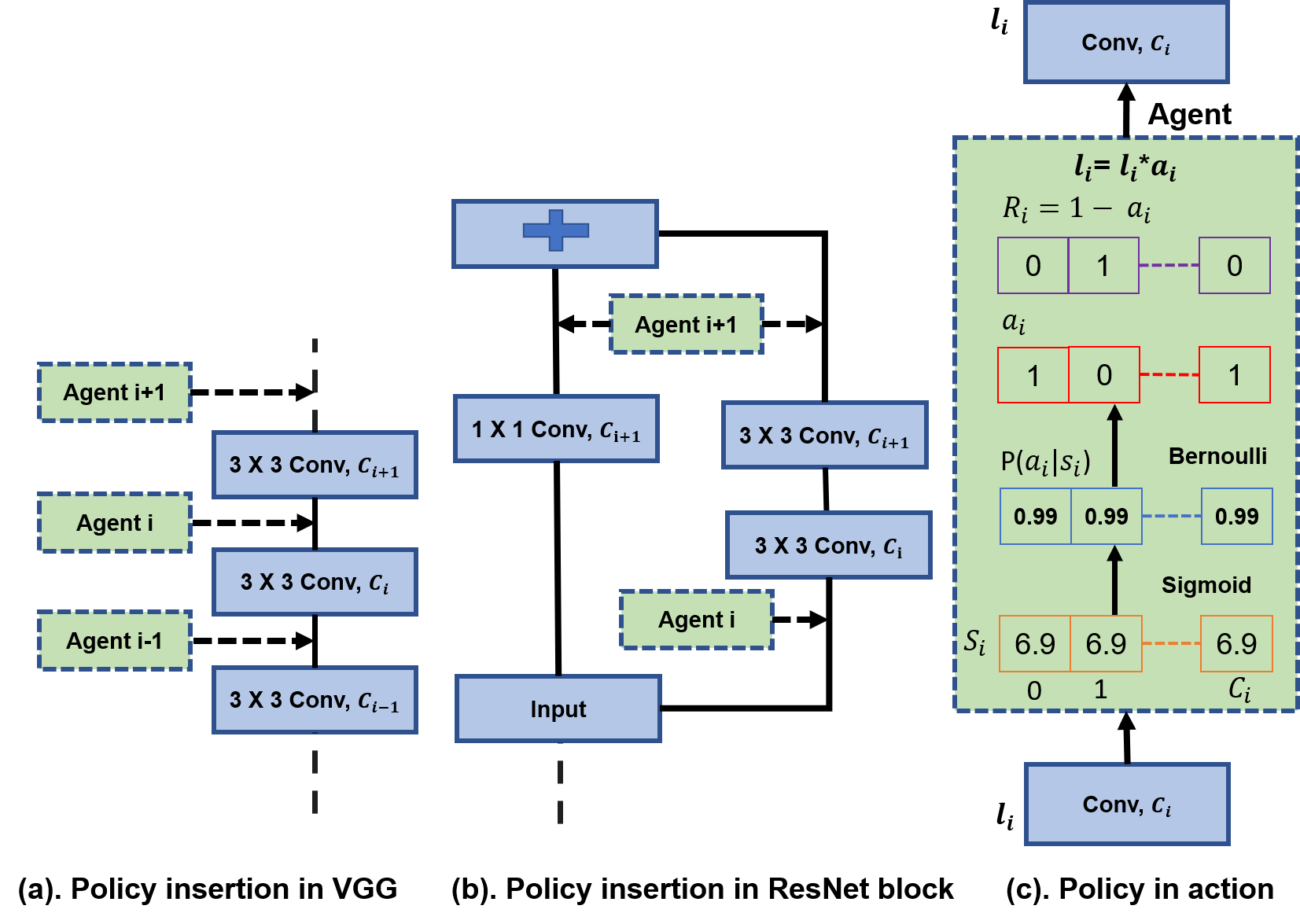} 
\caption{\textbf{DECORE Policies} (a) The policy is inserted before each convolution layer of VGG. (b) For Resnet blocks, we insert the policy sequentially (agent i) and also in the parallel path (agent i+1). We keep the same policy (agent i+1) for the last layer of parallel paths so that it doesn't affect future operations. (c) Our agent samples actions $a_i$ and multiplies them with $l_i$ to drop channels and compute the reward $R_i$.}
\label{fig:POI}
\vspace{-0.5cm}
\end{figure}

\section{Approach}
\subsection{DNN compression in RL framework}
A Markov Decision Process (MDP) provides a mathematical framework to model decision making in stochastic environments. At each time step, $t$, the agent is at state $s_t$, where it can choose any available action $a_t$ and move to the next state $s_{t+1}$, while accruing reward $R(s_t, a_t)$. The probability of moving to a new state $s_{t+1}$, given $s_t$ and $a_t$, is determined by the state transition probability $P(s_{t+1}| s_t,a_t)$. The objective of RL is to find a policy $\pi(a_t | s_t)$, which specifies the action an agent will take at $s_t$ to maximize some notion of cumulative reward~\cite{sutton2018reinforcement}. 

We can therefore frame the network compression process as an RL problem. At each layer $l_{i}$ with $C_{i}$ channels, we assign dedicated agents to all channels in that layer. Each agent decides to drop or keep its channel and subsequently receives reward $R_{i,j}$ (channel $j$ at layer $i$). The neural network does inference with channels which are not dropped and reaches the next state/layer $l_{i+1}$. Dropped channels can then be pruned from the network to create a compressed network. We train agents with a reward that incentivizes dropping channels at each layer $l_i$ without affecting network performance significantly. This creates a multi-agent RL framework. As compared to traditional MDP, the number of time steps ($t$) for each agent in this framework is just 1 as at each forward pass they take a decision to keep or drop channel only once and receive a reward based on that decision. 

\subsection{State Representation}
We model each layer of a neural network with a channel vector $s_{i} \in \mathbb{R}^{C_{i}}$, where $C_{i}$ is the number of channels in layer $i$. Each element $w_{j}$ of $s_{i}$ represents the weight (importance) of channel $j$ in layer $i$. A high value of $w_{j}$ means that channel $j$ is very important and dropping it would affect the network accuracy significantly. We initialize all channel vectors $s_{i}$ with the same value and our algorithm learns to increase the weight values of channels that lead to high network accuracy and reduce the weight of those that do not. After training, low weight channels are removed from the network, resulting in a compressed model. 

Our state representation does not take into account any other layer dependent variables, such as hidden values, height, width, number of input/output channels, kernel size, etc. \cite{ashok2018nn, AMC_He_2018_ECCV}, which makes this approach very general and applicable to a large or small number of channels or layers if needed. For example, this approach could be used to find important input channels for network training \cite{Xu_2020_CVPR} or removing channels only from early layers which have a high memory consumption and DRAM access time \cite{ferdman-micro-fused-layer-cnn-accelerators}. 

\subsection{Policy Representation}\label{PR}
The agent makes the decision at each layer $i$ to maximize the reward, based on a simple policy:
\begin{gather*}\label{eq:1}
\scalebox{.9}{$p_{j} = \frac{1}{1 + e^{-w_{j}}}$}\\
\scalebox{.9}{$\pi_{j} = \left\{
        \begin{array}{cc}
                1 & \mathrm{with\ } p_{j} \\
                0 & \mathrm{with\ } 1 - p_{j} \\
        \end{array} 
    \right.$}\\
\scalebox{.9}{$
a_{i}= \pi_{i}= \{\pi_{0}, \pi_{1}, ..., \pi_{C_{i}}\}$}
\end{gather*}
We convert the weights into probabilities via a sigmoid function and then choose which channels to keep at layer $i$ by taking independent Bernoulli samples. Sample values are either 0 or 1, where 0 means dropping the channel and 1 keeping it. During training, we multiply each layer's activation $l_{i}$ with $a_{i}$ to drop or keep channels. This is similar to dropout~\cite{JMLR:v15:srivastava14a} but here the probability of dropping the channel depends on the agent weight. There is a single parameter (weight) for each agent. This is in contrast to policies parameterized by neural networks in deep RL~\cite{mnih2013playing, mnih2016asynchronous}, which makes our approach simple and fast as we don't need to train dedicated neural networks to find good policies.

During training, we initialize the state vectors with a value of 6.9, which corresponds to a value of 0.99 after applying the sigmoid function. We assume initially that all channels in a pre-trained network are important and thus the initial probability of keeping a channel should be high. Each channel has a 99\% chance of being selected during sampling and 1\% chance not, which encourages little exploration and assists policy optimization. Figure~\ref{fig:POI} shows our policy in action for a layer $l_{i}$. 

We insert policies only before convolution or linear layers of the network. We avoid inserting policies in batch normalization or depth wise convolution layers, as they operate on individual channels and the decision to drop channels could be affected by the next layer's policy. Our policy insertion also depends on the network architecture. For example, ResNet or MobileNet architectures consist of residual blocks where computation happens in parallel paths and they are merged with a channel-wise addition. For these blocks, we keep the same policy at the end of parallel paths so that the same channels are dropped before merging. Figure~\ref{fig:POI} shows the policy insertion for VGG~\cite{simonyan2014} and ResNet~\cite{resnet2015} architectures. We use a similar approach to insert policies in computation layers of diverse architectures like GoogleLeNet~\cite{googlelenet} and DenseNet~\cite{densenet}. 

\subsection{Reward}\label{RT}
The objective of our modeling is to compress the network while maintaining predictive accuracy. We combine two different rewards to balance these considerations.

At each layer $i$, dropped channels represents the compression achieved at that layer. We compute compression reward $R_{i,C}$ at layer $i$ by summing all the dropped channels in that layer. We compute the number of dropped channels using the layer's action vector $a_i$: 
\begin{gather*}\label{eq:2}
    \scalebox{0.9}{$R_{i,C} = \sum_{j=1}^{C_{i}} 1 -a_{i,j}$}
\end{gather*}
After taking a series of actions at each layer, the network makes a prediction, $\hat{y}$, using only information from preserved channels. Each correct prediction suggests that the agent has taken the right actions and thus should be rewarded. An incorrect prediction incurs a penalty, $\lambda$, which can be tuned to adjust the preference for compression or accuracy: 
\begin{gather*}\label{eq:3}
\scalebox{.9}{$R_{acc} = \left\{
        \begin{array}{cc}
                1 & \mathrm{if\ } \hat{y} = y \\
                -\lambda & else
        \end{array} 
    \right.$}
\end{gather*}
The penalty is a high negative value which forces the agent not to drop the channels if predictions are wrong. Our final reward at each layer combines these rewards via multiplication. The value is high only when both compression and accuracy rewards are high,
\begin{gather*}\label{eq:4}
    \scalebox{0.9}{$R_{i} = R_{i,C} * R_{acc}$}
\end{gather*}

\subsection{Optimization with Reinforcement Learning}
The agent must learn to take decisions to maximize the expected sum of rewards across each layer. We can estimate the expected reward using trajectories $\tau$ sampled from the policy. In our framework, the agent takes the decision only once so the trajectory has only one time step. For example, the trajectory for layer $i$ is $\tau_i = (s_i, a_i)$. The optimization objective in this case is:
\begin{gather*}\label{eq:5}
    \scalebox{0.9}{$w* = \arg\max_{w} J(w) = \sum_{i=1}^{L} \arg\max_{w} E_{\tau_i\sim\pi_w(\tau_i)} [R_{i}]$}
\end{gather*}
We use the REINFORCE policy gradient algorithm~\cite{williams1992simple, mnih2016asynchronous} to learn policy parameters $w$,
\begin{gather*}\label{eq:6}
     \scalebox{0.9}{$\bigtriangledown_{w}J(w) = \sum_{i=1}^{L} \bigtriangledown_{w} E_{\tau_i\sim\pi_w(\tau_i)} [ R_{i}]$}\\
     \scalebox{0.9}{$\bigtriangledown_{w}J(w) = \sum_{i=1}^{L} E_{\tau_i\sim\pi_w(\tau_i)} [\bigtriangledown_{w}\log \pi_{w}(a_{i}|s_{i}) R_{i}]$}\\
     \scalebox{0.9}{$\bigtriangledown_{w}J(w) = \frac{1}{N}\sum_{b=1}^{N} \sum_{b=1}^{L}[\bigtriangledown_{w}\log\pi_{w}(a_{b,i}|s_{b,i}) R_{b,i}]$}
\end{gather*}
where $N$ is the batch size used to estimate the gradient and $L$ is a depth of the network. The gradients are used to update the policy or weights at each layer. We train policies and neural networks jointly using their own loss functions. 

In traditional RL approaches the state representation is stochastic and the agent takes an action by looking at each state to learn policies. This is contrary to our approach as we try to learn state representation without requiring expensive policy optimization methods like Actor-Critic~\cite{konda2000ActorCritic} and PPO~\cite{DBLP:journals/corr/SchulmanWDRK17}. Given that each agent has only one parameter, the policy gradient approach learns fast and efficiently.  

\section{Experiments}\label{sec:Exps}
\subsection{Datasets and Model Architectures}
We evaluate our approach by measuring the accuracy, compression, and acceleration (FLOPs reduction) rates of standard deep networks on common image classification benchmark datasets, such as CIFAR10~\cite{cifar10} and ImageNet~\cite{imagenet}. To demonstrate effectiveness of our approach we experimented with multiple widely-used model architectures, including VGG~\cite{simonyan2014} with sequential layers, ResNet~\cite{resnet2015} with residual blocks, GoogleLeNet~\cite{googlelenet} with inception module, and DenseNet~\cite{densenet} with dense block. 
\subsection{Training Details}
\label{sec:TD}
We used PyTorch~\cite{PyTorch} to implement the proposed method but our approach is very flexible and can be integrated with any deep learning library like TensorFlow~\cite{AbaBar16Tensorflow}. To train policies, we used the ADAM optimizer~\cite{adam} with a batch size of 256 and a learning rate of 0.01. Training begins with fixed policy vectors of an initial value 6.9 (Section \ref{PR}), yet different subnetworks are generated in each forward pass due to Bernoulli sampling. For each correct prediction, the agent gets a positive reward, and for each wrong prediction, it is penalized. We found the values of the penalty hyperparameter $\lambda$ through a grid search, running each experiment for 10 epochs. The high penalty leads to better accuracy as the agent is more penalized for every wrong prediction. As we lower the penalty it tries to compress the network more without largely affecting accuracy. 

Additionally, we also fine-tune the network at each optimization step so that it may recover from drops in performance after pruning. Joint training of policies and the network is more computationally efficient and sample efficient than approaches that separate learning from search \cite{tan2019efficientnet, lin2020hrank, li2016pruning}. On the CIFAR dataset, all models are trained for 300 epochs. We stopped policy training after 260 epochs and dropped all the channels with a probability of less than 50\%. We continued fine-tuning the remaining subnetwork in the next 40 epochs to improve accuracy. Finally, we removed the dropped channels from the network to get the compressed model. On the ImageNet dataset, models were trained for just 30 epochs. We stopped policy training after 20 epochs and fine-tune in the remaining epochs. 

\subsection{Results and Analysis}
\subsubsection{Results on CIFAR-10}\label{MN}
We evaluated the performance of the proposed approach on CIFAR-10 with four diverse networks, VGGNet-16/19, DenseNet-40, GoogleLeNet, and ResNet-56/110.  We used VGGNet with the variation of the original VGG-16 for CIFAR-10 similar to~\cite{li2016pruning, lin2019gal, lin2020hrank}. For GoogLeNet, the final output layer was changed to have 10 classes. DenseNet-40 has 40 layers with a growth rate of 12. For a fair comparison, we used the same trained networks used by~\cite{lin2019gal, lin2020hrank}. For all our experiments, we tried to adjust the penalty parameter ($\lambda$) to achieve similar accuracy as reported state-of-the-art results and compared the parameters/FLOPs reduction rates.

\textbf{VGGNet:} Table\ref{tab:vgg} shows that with a higher penalty ($\lambda=500$), our approach \textbf{DECORE-500} was able to achieve 0.15\% higher accuracy than the baseline model with a 63\% parameter and 35\% FLOPs reduction rate, which shows that this network was over parameterized for this problem and DECORE can be used as a regularizer. 

As compared to approaches that prune networks using network statistics~\cite{li2016pruning, lin2020hrank}, our approach provides higher compression and acceleration (FLOPs reduction) rates at similar or higher accuracy. As compared to L1 pruning~\cite{li2016pruning}, our approach DECORE-200 was able to get a 25\% higher compression rate (89\% vs 64\%) and a 30\% higher FLOPs reduction rate (34.3\% vs 64.8\%) at 0.16\% higher accuracy (93.56\% vs 93.40\%). Similarly, as compared to HRANK~\cite{lin2020hrank}, our approach provides better results at various compression rates. For example, with a high penalty, DECORE-200 was able to achieve 6.1\% higher compression and 11.3\% higher FLOPs reduction rates at 0.13\% higher accuracy (93.56\% vs 93.43) compared to HRANK-1. Similarly, with a low penalty, DECORE-50 was able to achieve 6.3\% higher compression and 11.8\% higher FLOPs reduction rates at 0.45\% higher accuracy (91.68\% vs 91.23\%) as compared to HRANK-3. Moreover, HRANK~\cite{lin2020hrank} manually sets different compression rates at each layer for higher compression or accuracy, and our method automatically (just by changing penalty $\lambda$) finds an optimal setting and generates better results overall. For the training process, HRANK trains each layer for 30 epochs (480 epochs for 16 layers), and the training time increases as the number of layers increases, while our approach only needs to train for 300 epochs for all settings.

\begin{table}
  \centering
  \resizebox{1.0\columnwidth}{!}{
  \begin{tabular}{lccc}
    \toprule
    Model&Top-1(\%)&Params(PR)&FLOPs(PR)\\
    \midrule
    VGG16            & 93.96  & 14.98M(0.0\%) & 313.73M(0.0\%)\\
    \textbf{DECORE-500}        & 94.02 & 5.54M(63.0\%) & 203.08M(35.3\%)\\ 
    L1~\cite{li2016pruning}           & 93.40 & 5.40M(64.0\%)& 206.00M(34.3\%) \\
    SSS$^*$~\cite{huang2018data}          & 93.02 & 3.93M(73.8\%)& 183.13M(41.6\%)  \\
    Zhao et al.~\cite{zhao2019variational}   & 93.18 & 3.92M(73.3\%)& 190.00M(39.1\%) \\
    HRank-1~\cite{lin2020hrank}           & 93.43 & 2.51M(82.9\%) & 145.61M(53.5\%) \\
    \textbf{DECORE-200}   & 93.56 & 1.66M(89.0\%)& 110.51M(64.8\%)\\
    GAL-0.05~\cite{lin2019gal}     & 92.03 & 3.36M(77.6\%)& 189.49M(39.6\%) \\
    HRank-2~\cite{lin2020hrank}           &92.34  & 2.64M(82.1\%)& 108.61M(65.3\%) \\
    \textbf{DECORE-100}   & 92.44 & 0.51M(96.6\%)& 51.20M(81.5\%)\\
    GAL-0.1~\cite{lin2019gal}      &90.73  & 2.67M(82.2\%)& 171.89M(45.2\%) \\
    HRank-3~\cite{lin2020hrank}           &91.23  & 1.78M(92.0\%)& 73.70M(76.5\%)  \\
    \textbf{DECORE-50}   & 91.68 & 0.26M(98.3\%)& 36.85M(88.3\%)\\
    \midrule
    VGG19            & 93.76  & 20.0M(0.0\%) & 398.44M(0.0\%)\\
    N2N\cite{ashok2018nn}&91.64&1.0M(95.0\%)&-\\
    \textbf{DECORE-40}&91.65&0.3M(98.5\%)&43.9M(89.0\%)\\
    \bottomrule
  \end{tabular}
  }
  \caption{Pruning results of VGGNet on CIFAR10. In all tables and figures, PR represents pruned rate, GAL-$X$ refers to GAL with sparsity factor $X$, DECORE-$\lambda$ refers to DECORE with penalty $\lambda$, and HRANK-$N$ refers to HRANK with a compression rate setting $N$ and M/B means millions/billions. For a fair comparison, we provide multiple results of HRANK, GAL, and others at various compression rates. $SSS^*$ shows the results standardized by HRANK and GAL.}
  \label{tab:vgg}
  \vspace{-0.5cm}
\end{table}

Our approach also finds better results than existing methods which learn channel importance\cite{zhao2019variational, huang2018data}. For instance, as compared to SSS\cite{huang2018data}, DECORE-200 achieved 15.2\% higher compression and 23.2\% higher FLOPs reduction rates at 0.54\% higher accuracy (93.56\% vs 93.02\%). As compared to Zhao et al.\cite{zhao2019variational}, DECORE-200 achieved 15.7\% higher compression and 25.7\% higher FLOPs reduction rates at 0.38\% higher accuracy (93.56\% vs 93.18\%). 

Compared to architecture search approaches such as~\cite{lin2019gal}, DECORE-100 provided 19\% higher compression and 41.9\% FLOPs reduction rates at 0.41\% higher accuracy (GAL-0.05). For high compression (low penalty), DECORE-50 was able to achieve 16.1\% high compression and 43.1\% FLOPs reduction rates at 0.95\% higher accuracy as compared to GAL-0.1. As compared to RL based architecture search method N2N~\cite{ashok2018nn}, DECORE achieves 3.5\% higher compression on the VGG-19 network at similar accuracy. N2N~\cite{ashok2018nn} first searches for compressed network architecture then fine-tunes the winner architecture to gain higher accuracy. N2N~\cite{ashok2018nn} training can take more than 2500 epochs to get the best architecture. DECORE achieves similar accuracy and higher compression with just 300 epochs ($<9\%$).     

\begin{table}
  \small
  \centering
  \resizebox{1.0\columnwidth}{!}{
  \begin{tabular}{lccc}
    \toprule
    Model&Top-1(\%)&Params(PR)&FLOPs(PR)\\
    \midrule
    GoogLeNet & 95.05 & 6.15M(0.0\%)&1.52B(0.0\%)  \\
    \textbf{DECORE-500} & 95.20 & 4.73M(23.0\%)& 1.22B(19.8\%)\\
    L1$^*$~\cite{li2016pruning}   & 94.54 & 3.51M(42.9\%) &1.02B(32.9\%) \\
    HRank-1~\cite{lin2020hrank}     &94.53  & 2.74M(55.4\%)&0.69B(54.9\%) \\
    \textbf{DECORE-200} & 94.51 & 1.17M(80.9\%)& 0.33B(78.5\%)\\
    GAL-ApoZ$^*$~\cite{Hu2016arxiv}&92.11 & 2.85M(53.7\%)&0.76B(50.0\%)\\
    GAL-0.05~\cite{lin2019gal}&93.93 & 3.12M(49.3\%)&0.94B(38.2\%)\\
    HRank-2~\cite{lin2020hrank}     &94.07 & 1.86M(69.8\%)&0.45B(70.4\%)\\
    \textbf{DECORE-175} & 94.33 & 0.86M(86.1\%)& 0.23B(84.7\%)\\
    \bottomrule
  \end{tabular}
  }
  \caption{Pruning results of GoogleLeNet on CIFAR10. L1$^*$ and GAL-ApoZ$^*$ are results standardized by HRANK and GAL}
  \label{tab:googleLeNet}
  \vspace{-0.0cm}
\end{table}

\textbf{GoogleLeNet:} Table~\ref{tab:googleLeNet} shows at higher penalty DECORE-500 was able to find 0.15\% higher accuracy than the baseline model with 23\% compression and 19.8\% Flops reduction rates. DECORE-200 was able to find higher compression (38\% vs L1, 25.5\% vs HRANK-1) and lower FLOPs (45.6\% vs L1, 23.6\% vs HRANK-1) at similar accuracy, compared to approaches using network statistics, such as L1\cite{li2016pruning} and HRANK\cite{lin2020hrank}. At a low penalty setting, DECORE-175 provided 16.3\% higher compression and 14.3\% FLOPs reduction as compared to HRANK-2. 

Compared to architecture search methods DECORE-175 provided high compression (36.8\% vs GAL-0.05 and 32.4\% vs GAL-ApoZ), lower FLOPs (46.5\% vs GAL-0.05 and 34.7\% vs GAL-ApoZ), and higher accuracy(0.4\% vs GAL-0.05 and 2.22\% vs GAL-ApoZ). It shows that DECORE can be applied to inception modules to achieve high performance as compared to state-of-the-art approaches.   
\begin{table}
  \small
  \centering
  \resizebox{1.0\columnwidth}{!}{
  \begin{tabular}{llcc}
    \toprule
    Model&Top-1(\%)&Params(PR)&FLOPs(PR)\\
    \midrule
   DenseNet-40 & 94.81 & 1.04M(0.0\%)& 282.92M(0.0\%) \\
   \textbf{DECORE-175} & 94.85  & 0.83M(20.7\%)& 228.96M(19.1\%)\\
   Liu et al.-40\%~\cite{liu2017learning} & 94.81 &0.66M(36.5\%)&190.00M(32.8\%) \\
   GAL-0.01~\cite{lin2019gal}&94.29 & 0.67M(35.6\%)&182.92M(35.3\%)\\
   HRank-1~\cite{lin2020hrank}&94.24& 0.66M(36.5\%)& 167.41M(40.8\%)\\
   \textbf{DECORE-115} & 94.59 & 0.56M(46.0\%)& 171.36M(39.4\%)\\
   Zhao et al.~\cite{zhao2019variational}& 93.16 & 0.42M(59.7\%)&156.00M(44.8\%)\\
   GAL-0.05~\cite{lin2019gal}&93.53 & 0.45M(56.7\%)&128.11M(54.7\%)\\
  HRank-2~\cite{lin2020hrank}&93.68 & 0.48M(53.8\%)&110.15M(61.0\%)\\
  \textbf{DECORE-70} & 94.04 & 0.37M(65.0\%)& 128.13M(54.7\%)\\
  \bottomrule
  \end{tabular}
  }
  \caption{Pruning results of DenseNet on CIFAR10.  Liu et al.-$\alpha$\% means $\alpha$ percentage of parameters are pruned.}
  \label{tab:densenet}
  \vspace{-0.5cm}
\end{table}

\textbf{DenseNet-40:}  DenseNet is a sophisticated architecture with channel concat (a growth rate of 12) at each layer, and channels at one layer are used in the following layers. Dropping even a single channel in this architecture means dropping this channel from all the following layers. Table~\ref{tab:densenet} shows DECORE results for the DenseNet architecture. With a higher penalty, DECORE-175 was able to achieve 20.7\% compression and 19.1\% Flops reduction rates at similar accuracy as the baseline model. With a slightly lower penalty, DECORE-115 was able to compress baseline by 46\% with 39.4\% FLOPs reduction with just a 0.22\% accuracy drop, better results compared to other techniques. For higher compression, DECORE-70 achieved 65.0\% compression and 54.7\% FLOPs reduction with just a 0.77\% drop in accuracy. These results demonstrate that our approach can be applied to complex architectures too.

\textbf{ResNet-56/110:} Table ~\ref{tab:ResNet} summarizes results on the ResNet-56/110 architecture. For ResNet-56, DECORE-200 provided 49\% compression and 49.9\% FLOPs reduction rates without any accuracy loss. At 50\% FLOPs constraints, RL based architecture search method AMC~\cite{AMC_He_2018_ECCV} finds the best network with 91.9\% accuracy which is 1.4\% lower than DECORE-200. In a high compression setting (low penalty), DECORE-55 compressed the network by 85.3\% and reduced FLOPs by 81.5\% with a 2.4\% accuracy drop. As shown in the table, ResNet-56 provided higher compression and lower FLOPs rates than all other approaches.  

For ResNet-110, DECORE-300 provided 64.8\% compression and 61.8\% FLOPs reduction without accuracy loss. DECORE-300 provided higher compression and lower FLOPs than all other approaches while achieving higher accuracy. DECORE-175 compressed the network even further to 79.63\% with 76.9\% FLOPs reduction and just a 0.79\% drop in accuracy.

\begin{table}
  \small
  \centering
  \caption{Pruning results of ResNet-56/110 on CIFAR10.}
  \vspace{-0.0cm}
  \resizebox{1.0\columnwidth}{!}{
  \begin{tabular}{llcc}
    \toprule
    Model&Top-1(\%)&Params(PR)&FLOPs(PR)\\
    \midrule
    ResNet-56 & 93.26 & 0.85M(0.0\%)&125.49M(0.0\%) \\
    \textbf{DECORE-450} & 93.34 & 0.64M(24.2\%)& 92.48M(26.3\%)\\
    L1~\cite{li2016pruning} & 93.06 &0.73M(14.1\%)& 90.90M(27.6\%) \\
    NISP ~\cite{yu2018nisp}&93.01&0.49M(42.4\%)& 81.00M(35.5\%) \\
    GAL-0.6~\cite{lin2019gal} & 92.98 & 0.75M(11.8\%)& 78.30M(37.6\%) \\
    HRank-1~\cite{lin2020hrank}& 93.17& 0.49M(42.4\%)& 62.72M(50.0\%)\\
    AMC~\cite{AMC_He_2018_ECCV}& 91.90& -& 62.72M(50.0\%)\\
    \textbf{DECORE-200} & 93.26 & 0.43M(49.0\%)& 62.93M(49.9\%)\\
    He et al.~\cite{He2017ICCV}& 90.80 & - & 62.00M(50.6\%) \\
    GAL-0.8~\cite{lin2019gal} & 90.36& 0.29M(65.9\%)& 49.99M(60.2\%)\\
    HRank-2~\cite{lin2020hrank}& 90.72& 0.27M(68.1\%)& 32.52M(74.1\%)\\
    \textbf{DECORE-55} & 90.85 & 0.13M(85.3\%)& 23.22M(81.5\%)\\
    \midrule
    ResNet-110 &93.50&1.72M(0.0\%)& 252.89M(0.0\%) \\
    \textbf{DECORE-500} & 93.88 & 1.11M(35.7\%)& 163.30M(35.4\%)\\
    L1~\cite{li2016pruning}&93.30& 1.16M(32.6\%)& 155.00M(38.7\%)\\
    GAL-0.5~\cite{lin2019gal}&92.55 & 0.95M(44.8\%)&130.20M(48.5\%)\\
    HRank-1~\cite{lin2020hrank} &93.36 & 0.70M(59.2\%)&105.70M(58.2\%)\\
    \textbf{DECORE-300} & 93.50 & 0.61M(64.8\%)& 96.66M(61.8\%)\\
    HRank-2~\cite{lin2020hrank} &92.65 &0.53M(68.7\%)&79.30M(68.6\%) \\
    \textbf{DECORE-175} & 92.71 & 0.35M(79.6\%)& 58.37M(76.9\%)\\
    \bottomrule
  \end{tabular}
  }
   \vspace{-0.5cm}
  \label{tab:ResNet}
\end{table}
\subsubsection{Results on ImageNet}\label{MN}
Table~\ref{tab:ImageNet} shows our performance with different penalty values on the challenging ImageNet dataset. DECORE-8 shows at higher penalty ($\lambda=8$), our model was able to achieve 11.0\% compression (25.50M to 22.69M) and 13.5\% (4.09B to 3.54B) FLOPs reduction at a 0.16\%/0.18\% Top-1/Top-5 accuracy improvement. 

With a slightly lower penalty ($\lambda=6$), DECORE-6 was able to achieve 74.56\%/92.18\% (1.57\%/0.68\% drop as compared to the baseline model) Top-1/Top-5 accuracy with the number of parameters reduced to 14.10M (44.74\% of baseline) and FLOPs reduced to 2.36B (42.30\% of baseline). Compared to approaches relying on network statistics such as ThiNet-70~\cite{thinet2017}, DECORE-6 achieved 2.54\% higher Top-1 accuracy and a 16.77\% higher compression rate with similar FLOPs. As compared to approaches that learn channel importance, such as SSS-32~\cite{huang2018data} and He et al.~\cite{He2017ICCV}, DECORE-6 provided higher accuracy, higher compression, and lower FLOPs. Furthermore, compared to architecture search methods such as GAL-0.5~\cite{lin2019gal}, DECORE-6 also achieved 33.49\% higher compression and 2.63\% (74.58\% vs 71.95\%) higher Top-1 accuracy. 

DECORE was able to find better performance in just 30 epochs which is much faster than architecture search approaches such as GAL (trains generator and discriminator in each batch) and pruning-based methods such as HRANK (90 epochs).

We found for a small dataset like CIFAR10, pre-trained models can often have very high accuracy and it becomes more challenging to prune these models without accuracy degradation. To meet this end the $\lambda$ parameter is set high so that we can achieve compression without any noticeable accuracy drop. On the other side, for more complex datasets like ImageNet, $\lambda$ has a lower value. 

\begin{table}
  \small
  \centering
    \caption{Pruning results of ResNet-50 on ImageNet. X-joint means jointly pruning channels and blocks.}
     \vspace{-0.0cm}
  \resizebox{1.0\columnwidth}{!}{
  \begin{tabular}{lllcc}
    \toprule
    Model&Top-1(\%)&Top-5(\%)&Params&FLOPs\\
    \midrule
    ResNet-50& 76.15 &92.87 &25.50M &4.09B \\
    \textbf{DECORE-8} &76.31&93.02& 22.69M& 3.54B\\
    SSS-32~\cite{huang2018data} & 74.18 & 91.91 &18.60M&2.82B  \\
    He et al.~\cite{He2017ICCV} & 72.30 & 90.80 &-&2.73B \\
    ThiNet-70~\cite{thinet2017} & 72.04& 90.67 &16.94M&2.44B \\
    GAL-0.5~\cite{lin2019gal} & 71.95 & 90.94 & 21.20M&2.33B\\
    \textbf{DECORE-6} &74.58&92.18& 14.10M& 2.36B\\
    GDP-0.6~\cite{gdp2018}&71.19 &90.71 &-&1.88B \\
    GDP-0.5~\cite{gdp2018}&69.58&90.14 &-&1.57B\\
    SSS-26~\cite{huang2018data}&71.82&90.79 &15.60M&2.33B \\
    GAL-1~\cite{lin2019gal}&69.88&89.75 & 14.67M&1.58B\\
    GAL-0.5-joint~\cite{lin2019gal} & 71.80 & 90.82 &19.31M&1.84B \\
    HRank-1~\cite{lin2020hrank} &71.98&91.01 & 13.77M&1.55B \\
    ThiNet-50~\cite{thinet2017} & 70.01& 90.02 &12.38M&1.71B \\
    \textbf{DECORE-5} & 72.06 &90.82& 8.87M& 1.60B\\
    ThiNet-30~\cite{thinet2017} & 68.42& 88.30 &8.66M&1.10B \\
    GAL-1-joint~\cite{lin2019gal}& 69.31& 89.12 &10.21M&1.11B \\
    HRank-2~\cite{lin2020hrank}& 69.10 & 89.58 & 8.27M& 0.98B \\
    \textbf{DECORE-4} & 69.71 &89.37& 6.12M& 1.19B\\
    \bottomrule
  \end{tabular}
  }
  \label{tab:ImageNet}
   \vspace{-0.5cm}
\end{table}

\begin{figure}
\centering
\includegraphics[width=10cm,height=5cm,keepaspectratio]{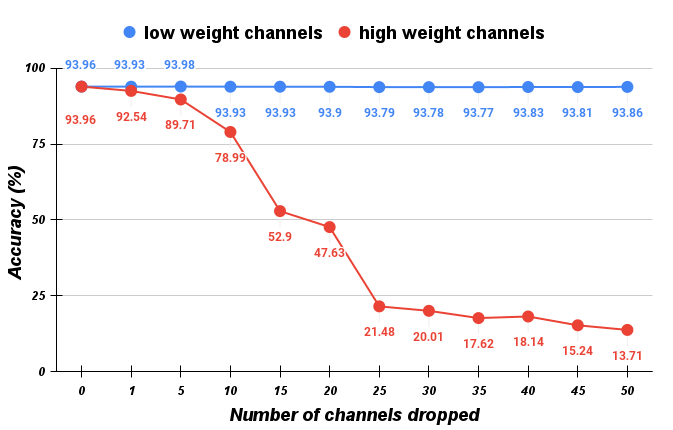} 
\caption{\textbf{DECORE finds important weights} After training VGG16 with DECORE, we remove high weight channels and observe significant drops in accuracy. On the flip side, removing low weight channels has a little impact on accuracy, and thus they can be dropped to compress the model. After dropping just 1\% (50) of highest weights channels accuracy drops significantly to 13\%, which highlights the large impact of those channels.}
\label{fig:channel drop with weights}
\vspace{-0.7cm}
\end{figure}

\section{Analysis: Does DECORE find important channels?}
\label{sec:analysis}
To investigate the effect of learned policies on network compression, we reviewed policies for the VGG16 network trained on CIFAR10 for 200 epochs with 4,736 agents (just convolution and linear layers). We didn't train the network jointly for this experiment so that the analysis of learned policies is not mixed with network update.

After training the agents, we remove channels with high weight values and observe how much accuracy drops. If removing high weight channels impacts network accuracy significantly, it suggests that the RL algorithm has correctly identified the most important channels in the network. Figure~\ref{fig:channel drop with weights} shows the effect on network accuracy after dropping low and high weight channels at each layer. The pre-trained network (without any channel drop) has an accuracy of 93.96\%. If we just remove the highest weight channel (1 channel out of 4736), there is a large accuracy drop of 1.42\% (92.54\% vs 93.96\% ). This decrease is significant, given that dropping this channel only compresses the network by 0.004\%. It suggests that DECORE is able to learn the most important channels, as dropping even one of them will impact accuracy significantly. In contrast, if we drop the lowest weight channel, the accuracy drop is negligible. If we drop 20 highest weight channels the network losses half of its accuracy (47.63\%), and after dropping top 50 channels the accuracy remains only 13\% which is almost the same as a random network. Dropping 50 of the lowest weight channels, however, does not impact accuracy much. This provides evidence that DECORE is able to identify channels with the least and most importance. 

\section{DECORE for compressed Network Architecture Search}

In real-world scenarios, there are often stringent limits on resources available. Although we can not directly optimize the network with hard constraints, we can tune our penalty reward to ensure trained models fall within those ranges. 
In section~\ref{sec:analysis}, we found DECORE learns weights for each channel in the network and the weight value shows the importance of that channel in the network. We also noticed that dropping low weight channels doesn't really impact the accuracy much. We can utilize this to sort all agents weight values and drop the lowest weight channels till the constraint is met. This approach can be used to meet different types of requirements, eliminating need to train networks with hard constraints. In this experiment, we explore the effect of hard constraints on performance. We also provide our findings on the MobileNetV2\cite{MobileNetv2} architecture which is hand designed for real world applications. It has 2.28M parameters which take 8.7MB of on-device memory and do 86.72M FLOPs during the inference on the CIFAR-10 dataset. Our baseline MobileNetV2 model has 92.58\% Top-1 accuracy for CIFAR-10.

In all experiments, we learned the policies for 200 epochs and then fine-tuned the compressed network for 60 epochs. Similar to Section~\ref{sec:analysis} we didn't train the network jointly while learning policies. 
\begin{table}
 \centering
 \caption{Top-1 accuracy with Size/FLOPs constraint on CIFAR}
 \label{tba1:Acc}
\resizebox{1.0\columnwidth}{!}{%
 \begin{tabular}{l|ccc|ccc}
    \toprule
     &&Size(MB)&&&FLOPs(M)&\\
     \midrule
      Model(Top-1\%)&0.5MB&1MB&5MB&25M&50M&100M\\
    \midrule
    VGG16&89.74&91.30&93.05&88.32&91.54&93.09\\
    ResNet-56&91.52&92.56&93.26&91.59&92.60&93.31\\
    MobileNetV2&90.97&91.95&92.58&92.01&92.31&92.58\\
    \bottomrule
 \end{tabular}
 }
 \vspace{-0.5cm}
\end{table}

Table~\ref{tba1:Acc} shows the performance drop of different models on the CIFAR10 dataset with different size constraints (in terms of storage memory and FLOPs). As explained above, we dropped the low weight channels till the constraint is satisfied. With a storage constraint of just 1MB, there is a maximum of 1\% performance drop across all models. When we decrease the size limit to 0.5MB, there is a negligible performance difference from the 1MB results. This illustrates that our technique is robust to varying memory constraints.

In addition to size constraints, we explore constraints on the number of FLOPs in a compressed network. FLOPs are highly correlated with latency on CPU-only devices and thus the results in Table~\ref{tba1:Acc} reflect performance with potential latency constraints. With a very low 50M FLOP limit, we observed a maximum performance drop of 2.47\% across all models. These results suggest that our approach can flexibly handle various hard constraints on computation, and by proxy latency, without significant drops in performance. More stringent constraints (size $<0.01MB$ or FLOPs $<10M$) lead to degenerate architectures and joint training (Section~\ref{sec:TD}) is recommended for these scenarios.

DECORE identifies the importance of each channel in the network once and then remove low priority channels to obtain a target architecture based on constraints. We don't need to retrain the network again when the target criteria is changed, which is contrary to other methods ~\cite{AMC_He_2018_ECCV, ashok2018nn}.

\section{Conclusion}\label{sec:Conc}
We present DECORE as an efficient reinforcement learning (RL) approach to network compression and compressed architecture search. Through our experiments on various datasets and architectures, we found that the size of compressed DECORE models is typically 3x smaller than those of similar approaches at the same level of accuracy, while the policy gradient algorithm selects the most important channels for prediction with a significant impact on model accuracy. Moreover, DECORE can find a trade-off between accuracy and compression by tuning the penalty for misclassifications, allowing for more control over the compressed model structure. Our method is simple yet very effective, and thus shows great promise as a tool for network compression. Although we demonstrate results with one type of reward, exploring other reward functions that target different resource constraints is an interesting direction to explore. We also leave the evaluation of our approach on architectures for other visual reasoning tasks (e.g. object detection, semantic segmentation, etc.) for future work.
One benefit of DECORE is that it finds compressed networks without doing architecture search, which could potentially save a lot of energy by reducing GPU hours during training. 
The compressed neural networks have a lower computation cost and are more energy efficient, which could be suitable for online inference on edge devices. 
We have not noticed any obvious negative consequences of our work. However, future extensions may have to consider their impact carefully.

{\small
\bibliographystyle{ieee_fullname}
\bibliography{ms}
}

\end{document}